\documentclass[letterpaper, 10 pt, conference]{ieeeconf}  

\IEEEoverridecommandlockouts                              

\usepackage{graphicx}
\graphicspath{ {../common/} }
\usepackage{amsmath}
\usepackage{booktabs}
\usepackage{algorithm}
\usepackage{algorithmic}
\usepackage{multirow}
\usepackage{arydshln} 
\usepackage{subcaption}
\usepackage{url}

\usepackage{cite}
\usepackage[switch]{lineno}
\usepackage[hidelinks]{hyperref}

\title{\LARGE \bf
Shared Representation for 3D Pose Estimation, \\ Action Classification, and Progress Prediction from Tactile Signals
}

\author{
Isaac Han$^{1}$, Seoyoung Lee$^{1}$, Sangyeon Park$^{1}$, Ecehan Akan$^{1}$,  \\ Yiyue Luo$^{2}$, Joseph DelPreto$^{3}$, Kyung-Joong Kim$^{1}$\\
\\%
\textbf{
$^{1}$Gwangju Institute of Science and Technology (GIST)\quad
$^{2}$University of Washington\quad
$^{3}$MIT CSAIL
}
}

\begin{document}


\maketitle
\thispagestyle{empty}
\pagestyle{empty}

\begin{abstract}
Estimating human pose, classifying actions, and predicting movement progress are essential for human–robot interaction. While vision-based methods suffer from occlusion and privacy concerns in realistic environments, tactile sensing avoids these issues. However, prior tactile-based approaches handle each task separately, leading to suboptimal performance. In this study, we propose a \textbf{S}hared \textbf{CO}nvolutional \textbf{T}ransformer for \textbf{T}actile \textbf{I}nference (\textbf{SCOTTI}) that learns a shared representation to simultaneously address three separate prediction tasks: 3D human pose estimation, action class categorization, and action completion progress estimation. To the best of our knowledge, this is the first work to explore action progress prediction using foot tactile signals from custom wireless insole sensors. This unified approach leverages the mutual benefits of multi-task learning, enabling the model to achieve improved performance across all three tasks compared to learning them independently. Experimental results demonstrate that SCOTTI outperforms existing approaches across all three tasks. Additionally, we introduce a novel dataset collected from 15 participants performing various activities and exercises, with 7 hours of total duration, across eight different activities.

\end{abstract}


\section{Introduction}

\begin{figure}[t]
    \centerline{\includegraphics[width=90mm]{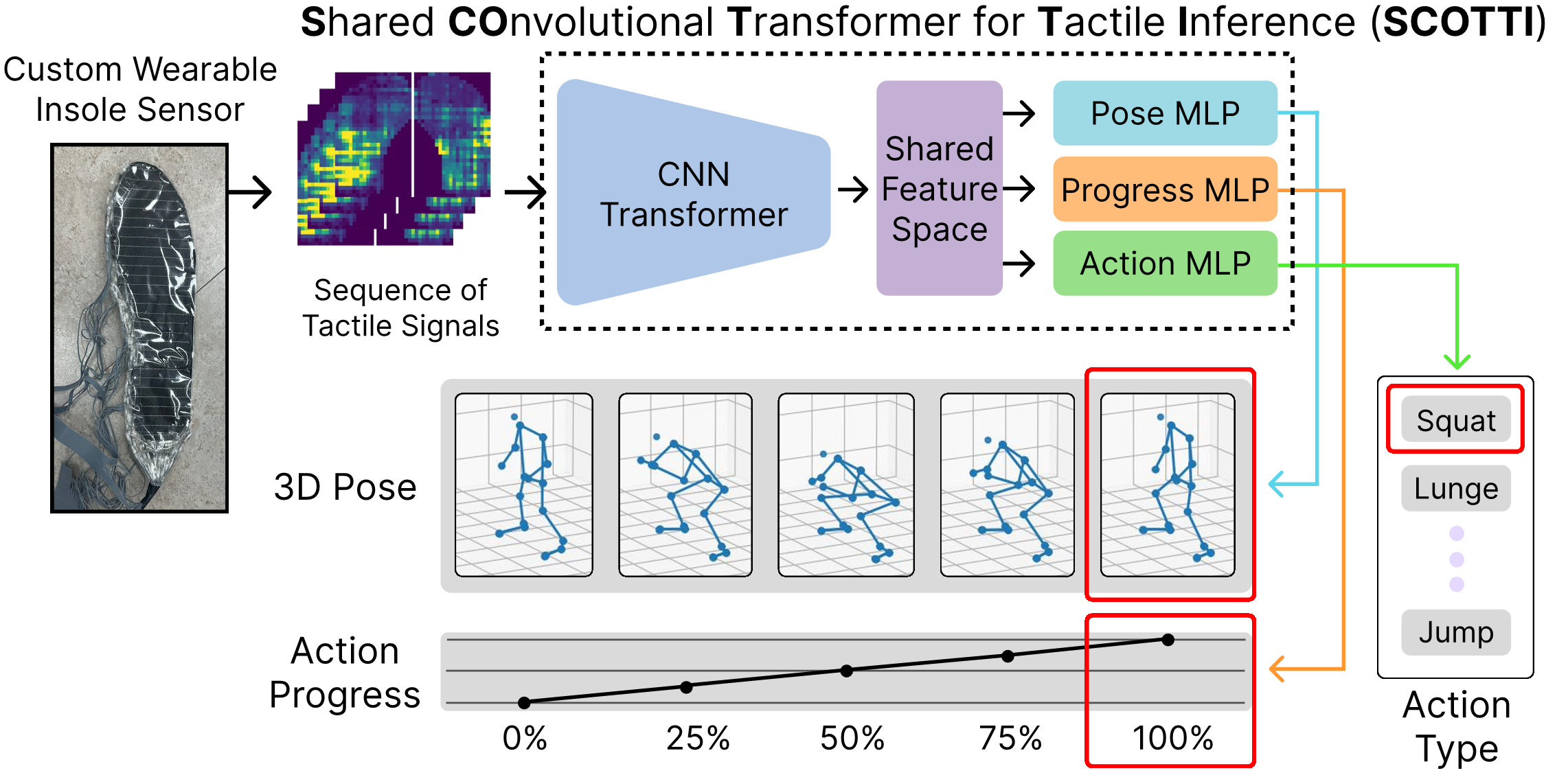}}
    \caption{\textbf{Overview of the proposed method.} The proposed model simultaneously performs 3D pose estimation, action progress prediction, and action classification by learning a shared representation from foot tactile signals collected with wearable insole sensors.}
\end{figure}

Estimating human pose, classifying actions, and predicting the progress of movements through sensor data are fundamental technology in human–robot interaction \cite{sheridan2016human}, with promising applications ranging from sports analysis to daily activity assistance. Particularly, such information serves as the foundation for understanding human intent and responding appropriately. This enables robots to support humans at the right moment, resulting in smoother and more intuitive human–robot interaction. 

Recent advances in deep learning and computer vision have enabled robots to infer human pose \cite{cao2019openpose}, classify actions \cite{host2022overview, sun2013active}, and estimate progress \cite{donahue2024learning, becattini2020done} directly from camera images.

However, in realistic and natural settings such as living and working environments, visual sensing suffers from occlusion and also raises privacy concerns.
By contrast, tactile sensing is inherently free from occlusion and privacy issues. Moreover, tactile sensors provide force and pressure information that is closely related to human posture but cannot be captured by visual sensors. This makes tactile sensing a strong alternative or complement to vision-based methods, supporting practical and more reliable pose, action, and progress estimation for robotics.

Although several studies have explored tactile-based human pose estimation \cite{wu2024soleposer, luo2021intelligent} and action classification \cite{lin2024jointly}, a key limitation is that these tasks are handled independently, which results in suboptimal performance. Yet, human pose, action type, and movement progress are all closely related aspects of human motion. 

Therefore, in this work, we hypothesize that learning the three tasks in a multi-task framework with shared representations will allow them to benefit from each other, thereby improving overall performance.


To effectively train and evaluate our model, we developed a custom high-resolution wearable insole sensor based on a prior work \cite{luo2021intelligent}. Our insole sensor is low-cost (approximately \$50 per unit) and supports wireless data transmission, eliminating the need for dedicated sensing infrastructure or workspace constraints. 
Using this new device, we collected more than 200,000 synchronized tactile and visual recordings, spanning over 7 hours, for ground-truth pose labeling from 15 participants performing a range of activities, including squatting, lunging, jumping, step-ups, and different walking styles.
These daily exercises are both natural and diverse, making them highly suitable for training models to estimate pose, action, and progress in everyday contexts, with potential extensions to exercise and sports training as well as elderly care. We plan to release the dataset and sensor design as an open-source.


To this end, we present the Shared COnvolutional Transformer for Tactile Inference (SCOTTI), a unified model that predicts 3D human poses, action classes, and action progress from tactile signals. By combining convolutional and transformer architectures, SCOTTI captures both spatial structure and temporal progression in tactile signals. SCOTTI outperforms prior approaches across all three tasks, demonstrating the benefits of multi-task learning on tactile data. Furthermore, by visualizing the learned feature space, we show that SCOTTI acquires highly meaningful shared representations of human movement.

Our main contributions are summarized as follows:

\begin{itemize} 
\item We propose a \textbf{S}hared \textbf{CO}nvolutional \textbf{T}ransformer for \textbf{T}actile \textbf{I}nference (\textbf{SCOTTI}) that simultaneously predicts 3D human poses, action classes, and action progress, leveraging mutual benefits across tasks.

\item To the best of our knowledge, this is the first work to address action progress prediction using tactile signals, extending the scope of tactile-based research.

\item We collected a large-scale dataset of over 7 hours of synchronized tactile and visual frames from 15 participants. The dataset includes various physical activities such as squats, lunges, jumps, step-ups, and multiple types of walking. This dataset serves as a valuable resource for advancing research in pose estimation, action classification, and progress prediction.

\item Evaluation results demonstrate the effectiveness of our SCOTTI model, outperforming existing methods in 3D pose estimation, action classification, and progress prediction using tactile signals.
\end{itemize}

\section{Related Works}

\subsection{3D Pose Estimation with Tactile Signals}

Recent methodologies employing images and videos to predict human poses have seen substantial advancements \cite{zheng2023deep}. Nonetheless, these camera-based methods encounter challenges such as occlusion and privacy concerns and further require dedicated physical space and infrastructure for camera installation, and are constrained to operate only within such instrumented environments. As an alternative, tactile sensor-based pose estimation approaches are proposed to overcome these limitations. 

Two primary types of tactile sensors have been used in these studies: carpet-type sensors and insole-type sensors. 

Carpet-type sensors were initially studied in scenarios where most of the human body is in contact with the floor. These studies primarily focused on data collected when people were lying on beds, targeting applications in the medical field \cite{clever2020bodies, casas2019patient, liu2024tagsleep3d}. Subsequent research shifted to more general scenarios, such as daily activities and exercises \cite{luo2021intelligent}. Using data extracted from carpet-type tactile sensors, they proposed CNN-based deep learning models to predict full-body 3D poses. Unlike our shared representation approach, they performed action classification by freezing the encoder of the trained model and adding a linear classifier on top for fine-tuning. Additionally, methods leveraging synthetic data from simulations to assist in 3D pose prediction, rather than collecting real tactile data, have also been proposed \cite{tripathi20233d, han2023groundlink}. However, a limitation of carpet-type sensors is their spatial constraint.

In contrast, insole-type wearable sensors provide better flexibility for movement without space constraints. Human pose estimation using insole sensors holds significant potential for various applications, as shoes are commonly worn in daily life. While some studies focused on lower-body pose prediction \cite{alemayoh2023neural, tam2019lower}, insole-type sensors are also capable of predicting full body 3D pose by utilizing high-resolution sensing \cite{luo2021learning}. The recent work achieved comparable performance to camera-based methods on 3D pose estimation by utilizing foot pressure and IMU together \cite{wu2024soleposer}.

These existing studies have proposed models that only focus on pose estimation, without performing additional tasks.

\subsection{Action Classification with Tactile Signals}

While there are relatively few studies on action classification using tactile signals, some notable research exists. A significant example is the spatio-temporal feature approach proposed by \cite{lin2024jointly}. Unlike earlier studies that relied on simple CNN networks \cite{sundaram2019learning}, they utilized transformers and introduced a pre-training task and embeddings capable of effectively processing both the spatial and temporal features of tactile data. Their method utilizes not only positional embeddings commonly used in transformer models, but also tubelet embeddings in video processing. Both of the two works are based on insole-type tactile sensor. 

\subsection{Action Progress Prediction and Stage Classification}

Progress prediction is an actively studied task in the visual domain. Early work utilized simple convolutional and recurrent networks to predict progress \cite{becattini2020done}. They labeled video clips by linearly assigning progress values ranging from 0 to 1, from the start to the end of the clip. A notable methodological improvement involved predicting progress by aligning images within videos. \cite{dwibedi2019temporal} proposed temporal cycle-consistency learning, which learns useful representations in a self-supervised manner by temporally aligning images across different videos. However, their work focused on action stage classification rather than predicting continuous progress. Subsequent studies enhanced this approach by incorporating global temporal alignment \cite{hadji2021representation} and video-level cues \cite{haresh2021learning}. Beyond action stage classification, \cite{donahue2024learning} addressed continuous progress prediction using self-supervised video alignment.

Intuitively, action progress prediction is related to action classification. \cite{shen2024progress} proposed a progress-aware online action segmentation approach for procedural task videos, where predicted progress was used to improve action classification. This mutual benefit has also been observed in other applications. For example, \cite{van2020multi} reported a performance boost when gesture recognition and progress prediction were simultaneously learned on surgical data. A similar approach was applied to action recognition and progress prediction for human-centric intelligent manufacturing data \cite{wang2024deep}.

Where these works are from the visual domain, there is related research in the tactile domain. \cite{luo2024tactile} explored stage classification for hand manipulation tasks, considering a limited number of stages (4-6) per action. However, the challenges of predefining a detailed dictionary of fine-grained phases, combined with the significant cost of annotating tactile frames for many action phases, make this approach impractical for real-world scenarios. Although they also proposed shared representations for tactile learning, their work was centered on hand actions rather than the more complex whole-body actions.

\section{Dataset}

In this section, we describe details of our dataset and tactile sensing hardware as well as labeling process for the ground truth 3D pose and progress labels.

\subsection{Tactile Sensing Hardware}

We build custom tactile sensing hardware to efficiently collect high-resolution data of foot pressure, which was adapted from prior sensor designs \cite{luo2021intelligent} into an insole form factor. Our insole sensor system incorporates a wireless piezoresistive pressure sensor insole, featuring over 500 sensors per foot, designed for high-resolution tactile data acquisition. Each insole consists of a grid formed by orthogonally aligned copper threads as electrodes on opposite sides of piezoresistive films. The intersections of these electrodes are the sensors, allowing for precise pressure mapping. Costing around \$50 per unit, these insoles are cost-effective and easy to manufacture, ideal for large-scale data collection. The tactile signals are read by a microcontroller board attached to the ankle and wirelessly transmitted at 16Hz to a paired board connected to a computer. The batteries and boards are attached to the ankle using elastic bands. 

\subsection{Data Collection}

To obtain ground-truth body pose during data collection, we used visual sensing; this visual input is only for labeling and is not required at deployment. We employed XRmocap \cite{xrmocap}, an open-source multi-view motion capture framework. Data was collected using six strategically positioned cameras to capture diverse viewing angles. We extracted 19 keypoints for each person, including head, neck, shoulders, elbows, wrists, hips, knees, ankles, heels, small toes, and big toes. 

Our high-resolution system accurately captures foot pressure, improving 3D human keypoints' prediction from tactile inputs alone. We have collected over 200,000 synchronized tactile and visual frames. This dataset was recorded from 15 participants, each performing eight distinct actions: squatting, lunging, step-up exercise, jumping, side walking, in-place walking, backward walking, and walking. These actions encompass a wide range of human movements, from common exercises such as squats and lunges to various walking patterns, providing rich information about human motion.

\subsection{Action Progress Labels}

A key contribution of our dataset is the provision of progress labels. In this section, we introduce how we labeled the progress for each data frame. 

In everyday exercises and movements such as squats, lunges, and walking, the definition of progress is intuitive and straightforward, as each action can be represented as a simple cycle — by start and end points for squats and lunges, or by a step cycle in walking. For example, in the case of squats, a single action is defined as the performer gradually lowering their body by pushing their hips back, returning to the starting position after reaching a sufficient depth. In other words, as the body descends from the starting position, the progress increases, and when the performer reaches the lowest position and returns to the starting position, the progress reaches 100\%.

This perspective can also be applied to define progress in other movements. For squats and lunges, we define a single action as starting from the initial posture, performing the movement, lowering the body, and then returning to the initial posture. For jumps, a single action is defined as the body leaving the ground and then landing back on it. A single action of step-up exercise is defined as stepping onto a platform with one foot, bringing the other foot onto the platform, and then stepping down. For various walking motions, a single action is defined as the process of taking a step forward with one foot.

Specifically, we define a progress variable \(p \in [0, 1]\) for a single repetition of an action using an arbitrary indicator value \(h\), which represents the progress-relevant measurement of the action. For example, \(h\) could be the vertical position of a joint, the angle of a limb, or any other meaningful measure. In case of squat, \(h\) could be \(z\)-position of middle hip. Assume that \(h\) moves periodically between a maximum value \(h_{\max}\) and a minimum value \(h_{\min}\). We mark:
\begin{itemize}
    \item \textbf{Start posture} (e.g., initial position): \(p = 0\), \(h = h_{\max}\), and \(t=t_{start}\).
    \item \textbf{Midpoint posture} (e.g., halfway through the action): \(p = 0.5\), \(h = h_{\min}\), and \(t=t_{mid}\).
    \item \textbf{End posture} (e.g., returning to the initial position): \(p = 1\), \(h = h_{\max}\), and \(t=t_{end}\).
\end{itemize}

We detected start, middle, and end points by detecting maximum and minimum peaks of the \(h\) values. The motion from \(h_{\max}\) to \(h_{\min}\) corresponds to progress increasing from \(0\) to \(0.5\), and the motion from \(h_{\min}\) back to \(h_{\max}\) corresponds to progress increasing from \(0.5\) to \(1\).

Let \(h(t)\) be the indicator value at time \(t\). Define:
\begin{itemize}
    \item \(h_{\max}\): the maximum observed value of \(h\) (start/end posture),
    \item \(h_{\min}\): the minimum observed value of \(h\) (midpoint posture).
\end{itemize}

We compute the progress \(p(t)\) as follows:

\begin{equation}
\small 
p(t) =
\begin{cases}
\displaystyle 
0.5 \times \frac{h_{\max} - h(t)}{h_{\max} - h_{\min}}, 
& \text{if } t_{start} \le t \le t_{mid},

\\[1.5em]
\displaystyle
0.5 + 0.5 \times \frac{h(t) - h_{\min}}{h_{\max} - h_{\min}}, 
& \text{if } t_{mid} < t \le t_{end}.
\end{cases}
\label{eq:progress}
\end{equation}

This method can be applied to a wide range of actions by appropriately choosing \(h\) and defining the mapping from \(h\)-values to the progress scale \([0,1]\). For lunge, squat, and step-up exercise, we define \(h\) as the \(z\)-position of the middle hip (\(M_{hip,z}\)). For jump, \(-M_{hip,z}\) is used. For walking and side-walking, \(h\) is the Euclidean distance between heels (\(d_{heel} = \|L_{heel} - R_{heel}\|\)). For in-place walking, \(h\) is the maximum \(z\)-position of the heels (\(\max(L_{heel,z}, R_{heel,z})\)), and for backward walking, \(h\) is the maximum change in heel positions over time (\(\max(d^{t-5}_{heel},\dots, d^t_{heel})\)), calculated over a window of 5 frames.

Note that $h$ is used solely for labeling and is not provided as input to the model. 
During training, only the progress value $p(t)$ is given as a label and is not used as an input feature, so the model must implicitly learn to predict $p(t)$. The tactile frames are the only inputs to the model.

\section{Method}

In this section, we introduce our SCOTTI model and describe details of network architecture and loss function. 

\subsection{Network Architecture}

We propose a SCOTTI to effectively extract and encode spatio-temporal features from tactile input data. The proposed model effectively combines the spatial feature extraction capabilities of CNNs with the sequential modeling strengths of transformers. The model predicts pose, action class, and action progress at time $t$, while receiving input data from the time range $t-1-T$ to $t-1$. The input data consists of a sequence of $T$ tactile signals from the left and right feet, structured as a sequence of 2D frames with dimensions $H \times 2W$ for a sliding window of size $T$. Each tactile frame for the left and right feet has a shape of $H \times W$.

To effectively handle spatial information, each frame is processed independently by the CNN feature extractor, which comprises two convolutional layers, each followed by ReLU activation functions and pooling layers. The output from the CNN feature extractor is flattened and passed through a fully connected layer to project it into an embedding space of dimension $E$. The embeddings of the $T$ frames are concatenated along the temporal dimension, forming a tensor of shape $T \times E$. A class token, initialized as a learnable parameter with dimensions $1 \times E$, is appended to the front of the sequence to allow the transformer encoder to learn a global representation of the tactile signals. Learnable positional encodings are added to the sequence embeddings to encode temporal information. The encoded sequence is then fed into the transformer encoder.

A weighted mean pooling is applied along the temporal dimension of the output of the transformer encoder, producing the final representation of the tactile signals. This shared representation is fed into three separate multi-layer perceptrons (MLPs) to predict pose, action class, and action progress, respectively.

\subsection{Loss Function}

For pose estimation, we use the sum of Mean Per Joint Position Error (MPJPE) and Mean Per Joint Angle Error (MPJAE). For action classification, we employ the standard cross-entropy loss, and for progress regression, we utilize the Mean Squared Error (MSE) loss. The total loss function is a weighted sum of the three losses. The total loss is defined as:
\begin{equation}
L_{\text{total}} = w_{\text{pose}} \, L_{\text{pose}} + w_{\text{action}} \, L_{\text{action}} + w_{\text{progress}} \, L_{\text{progress}},
\end{equation}

where \(L_{\text{pose}} = \text{MPJPE} + \text{MPJAE}\), and \(w_{\text{pose}}\), \(w_{\text{action}}\), \(w_{\text{progress}}\) are weighting coefficients for each term. Since the three tasks use different loss functions, the weights serve to normalize differences in scale, rather than to indicate task importance.

\section{Result}

\begin{table*}[t]
    \caption{\textbf{Comparison of SCOTTI with baselines.} Arrows indicate the desired direction for each metric, and bold numbers indicate the best performance.}
    \label{table:main}
    \centering
        \begin{tabular}{l cc c cc}
            \toprule
            \multirow{2}{*}{\textbf{Model}} & \textbf{Pose Estimation} & \textbf{Action Classification} & \multicolumn{2}{c}{\textbf{Progress Prediction}} \\
            \cmidrule(r){2-3} \cmidrule(l){4-4} \cmidrule(l){5-6}
            & \textbf{MPJPE (mm)} $\downarrow$ & \textbf{Accuracy (\%)} $\uparrow$ & \textbf{APP} $\uparrow$ & \textbf{MSE} $\downarrow$ \\
            \midrule
            Voxel-CNN & 104.91 & - & - & -  \\
            Sep-CNN & 100.57  & - & - & -  \\
            Res-CNN & 98.51  & - & - & -  \\
            GCN-Transformer & 117.81 & - & - & -  \\
            \midrule
            STAT & - & 84.56 & - & -  \\
            \midrule
            Random & - & - & 0.6904 &  0.1448  \\
            \midrule
            \midrule
            \textbf{SCOTTI (Ours)} & \textbf{96.63} & \textbf{90.06} & \textbf{0.8952} & \textbf{0.0223}  \\
            \bottomrule
        \end{tabular}
\end{table*}

\begin{table*}[t]
    \caption{\textbf{Comparison of SCOTTI with multi-task learning versions of baselines.} Arrows indicate the desired direction for each metric, and bold numbers indicate the best performance.}
    \label{table:multi}
    \centering
        \begin{tabular}{l cc c cc}
            \toprule
            \multirow{2}{*}{\textbf{Model}} & \textbf{Pose Estimation} & \textbf{Action Classification} & \multicolumn{2}{c}{\textbf{Progress Prediction}} \\
            \cmidrule(r){2-3} \cmidrule(l){4-4} \cmidrule(l){5-6}
            & \textbf{MPJPE (mm)} $\downarrow$ & \textbf{Accuracy (\%)} $\uparrow$ & \textbf{APP} $\uparrow$ & \textbf{MSE} $\downarrow$ \\
            \midrule
            Voxel-CNN & 103.55 & 80.38 & 0.8770 & 0.0288  \\
            Sep-CNN & 101.05 & 78.18 & 0.8668  & 0.0330 \\
            Res-CNN & 99.00 & 82.59 & 0.8778  & 0.0278\\
            GCN-Transformer & 124.27 & 61.02 & 0.8191 & 0.0537\\
            \midrule
            STAT & 99.61 & 86.36 & 0.8894 & 0.0242 \\
            \midrule
            \midrule
            \textbf{SCOTTI (Ours)} & \textbf{96.63} & \textbf{90.06} & \textbf{0.8952} & \textbf{0.0223} \\
            \bottomrule
        \end{tabular}
\end{table*}

\subsection{Model Training and Evaluation}

\textbf{Evaluation Method.} We divided the dataset of 15 participants into random splits of 10 for training and 5 for testing, ensuring that evaluation was always performed on participants unseen during training. 
To reduce dependency on a particular split, we created three train-test partitions, trained a separate model on each, and reported the final metrics as the average across these runs. All activities were trained together rather than separately by activity, and the same subject splits were consistently applied across pose estimation, action classification, and progress regression.\\

\noindent\textbf{Metrics.} For pose estimation, we evaluate using the standard MPJPE metric. For action classification, we report the classification accuracy across 8 action classes. For progress regression, we evaluate using the Mean Squared Error (MSE) and the Average Progress Precision (APP) metric proposed by \cite{becattini2020done}. Throughout the tables, we use arrows to indicate the desired direction for each metric (\(\downarrow\): lower is better, \(\uparrow\): higher is better), and highlight the best results in \textbf{bold}.

To compute APP, we first define a successful detection as a case where the absolute difference between the predicted progress \(\hat{z}_i\) and the ground-truth progress \(z_i\) is less than a progress margin \(m\):
\begin{equation}
\text{Success}_i = 
\begin{cases} 
1 & \text{if } |\hat{z}_i - z_i| < m, \\
0 & \text{otherwise}.
\end{cases}
\end{equation}
Based on this criterion, Progress Precision is calculated as the ratio of successful detections to the total number of samples. APP is then derived as the area under the Precision-Margin curve, where x axis is $m$ and y axis is Progress Precision.

\noindent\textbf{Baselines.} We demonstrate the superior performance of our approach by comparing our method with state-of-the-art architectures that proposed for pose estimation and action classification. These models are well-suited for predicting pose, action, and progress from tactile signals. They have demonstrated competitive performance on tactile sensing–based human pose estimation using specialized architectures, and several studies \cite{luo2021intelligent, luo2021learning} employed piezo-resistive sensors which is similar to ours, suggesting that the models are likely to work effectively with our data as well.

We considered four baselines for pose estimation, one baseline for action classification, and one baseline for progress prediction--a total of six baselines.

The baselines for pose estimation are as follows:
\begin{itemize} 
\item \textbf{Voxel-CNN} \cite{luo2021intelligent}: This model adopts a CNN to predict 3D heatmaps for each keypoint from tactile signals. The final pose prediction is generated by performing a soft argmax operation on the heatmaps.
\item \textbf{Sep-CNN} \cite{luo2021learning}: This model utilizes a two distinct CNN layers to separately process left and right foot tactile signals.
\item \textbf{Res-CNN} \cite{scott2020image}: PressNet is originally designed to predict foot pressure from human pose, we used an inversed version of PressNet to predict pose from foot pressure. This inversed version of PressNet already been used as a baseline in previous work \cite{wu2024soleposer}. The core architectural novelty of this model is residual connection, so we refer this model as Res-CNN
\item \textbf{GCN-Transformer} \cite{wu2024soleposer}: They utilizes Graph Convolutional Network (GCN) \cite{zhang2019graph} with transformer network to predict pose from foot pressure and IMUs, we used a pressure-only version of SoleFormer. Note that we made a minimal change from the original model by only removing the modules for IMUs. Also we adopted cyclic loss same as proposed in the paper.
\end{itemize}

The baselines for action classification are as follows:
\begin{itemize}
\item \textbf{STAT} \cite{lin2024jointly}: This utilizes temporal pretraining and spatio-temporal embeddings with VideoMAE approach \cite{tong2022videomae}, to perform action classification from foot tactile signals. We used codes provided by the authors.
\end{itemize}

Since there is no prior work on action progress prediction, we used random prediction as a baseline.

Additionally, we compared our model with a multi-task version of these baselines, incorporating the same MLP decoder structure as our model to facilitate multi-task learning across the three tasks.

\noindent\textbf{Implementation Details.} We used a window size of \( T=40 \) and an embedding dimension of \( E=512 \). The size of the tactile frames obtained from the left and right feet are \( H=32 \) and \( W=22 \), respectively. We used loss coefficient \( w_{\text{pose}}=0.01 \), \( w_{\text{action}}=1 \), and \( w_{\text{progress}}=1 \). The learning rate was set to $10^{-4}$, with a weight decay of $10^{-4}$. To ensure a reduction in loss, we employed learning rate scheduling and conducted training for 25 epochs for each train and test set pair. The model is optimized by Adam optimizer \cite{diederik2014adam}. We applied random shifting data augmentation to all baselines and our approach to prevent overfitting.

\begin{figure*}[t]
    \centerline{\includegraphics[width=180mm]{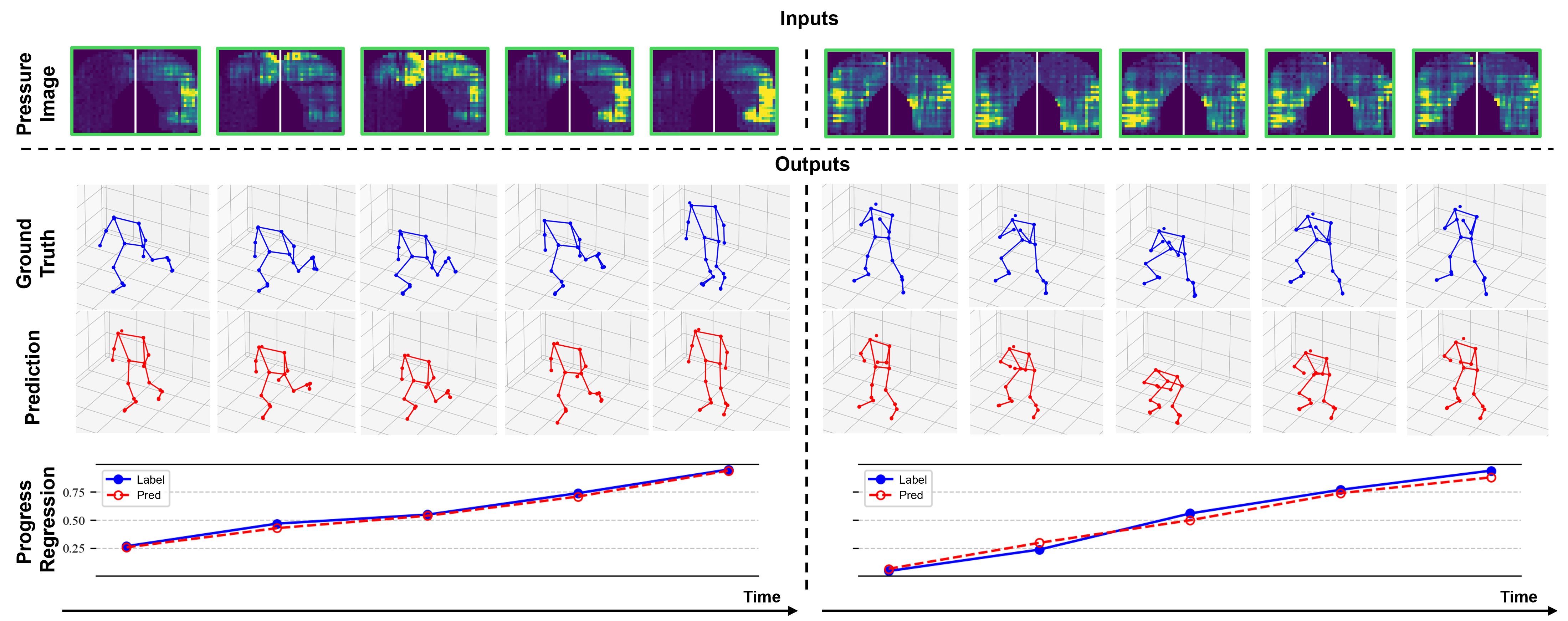}}
    
    \caption{\textbf{Qualitative result of pose estimation and progress prediction results for lunges (left) and squats (right).} The results demonstrate SCOTTI's ability to accurately predict both pose and progress for different activities.}
    \label{fig:qualitative}
\end{figure*}

\subsection{Overall Performance}

\begin{figure}[t]
    \centering
    \begin{subfigure}[t]{0.49\columnwidth} 
        \centering
        \includegraphics[width=\linewidth]{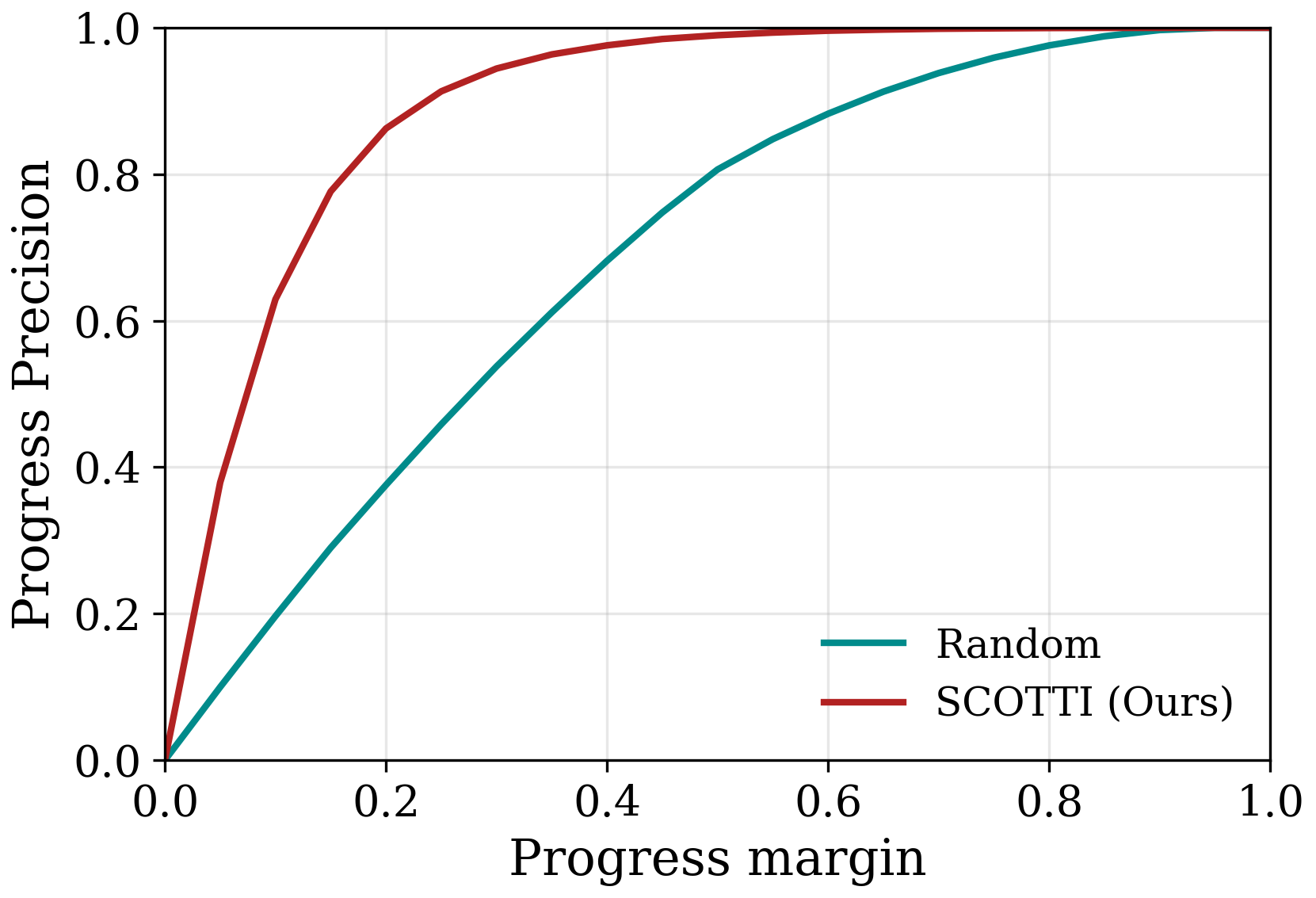}
    \end{subfigure}
    \begin{subfigure}[t]{0.49\columnwidth} 
        \centering
        \includegraphics[width=\linewidth]{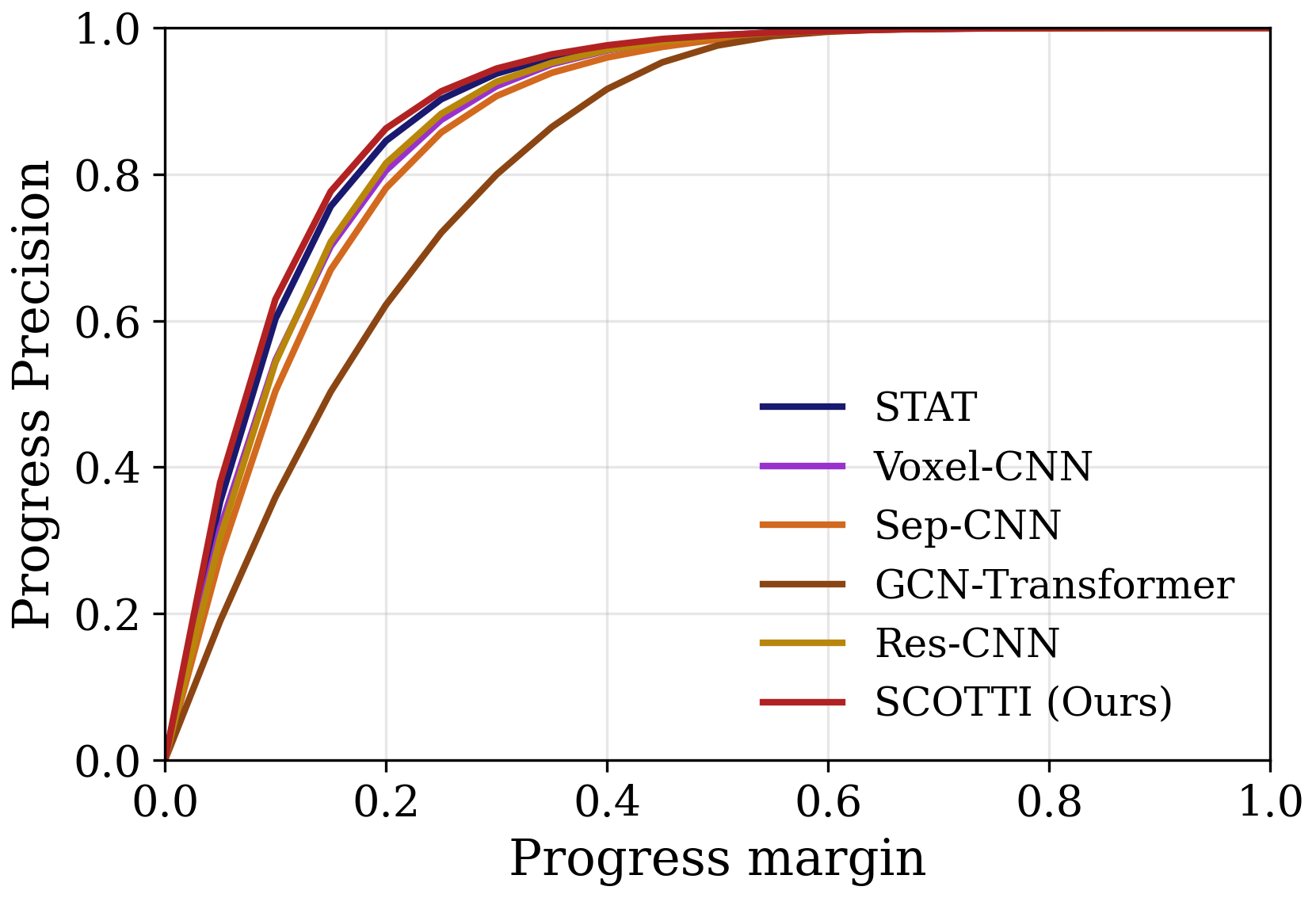}
    \end{subfigure}
    \caption{\textbf{Progress Precision-Margin (PM) curve.} PM-curve with random prediction baseline \textbf{(left)}. PM-curve with multi-task version of baselines \textbf{(right)}.}
    \label{fig:progress_curve}
\end{figure}

\begin{figure}[t]
    \centerline{\includegraphics[width=90mm]{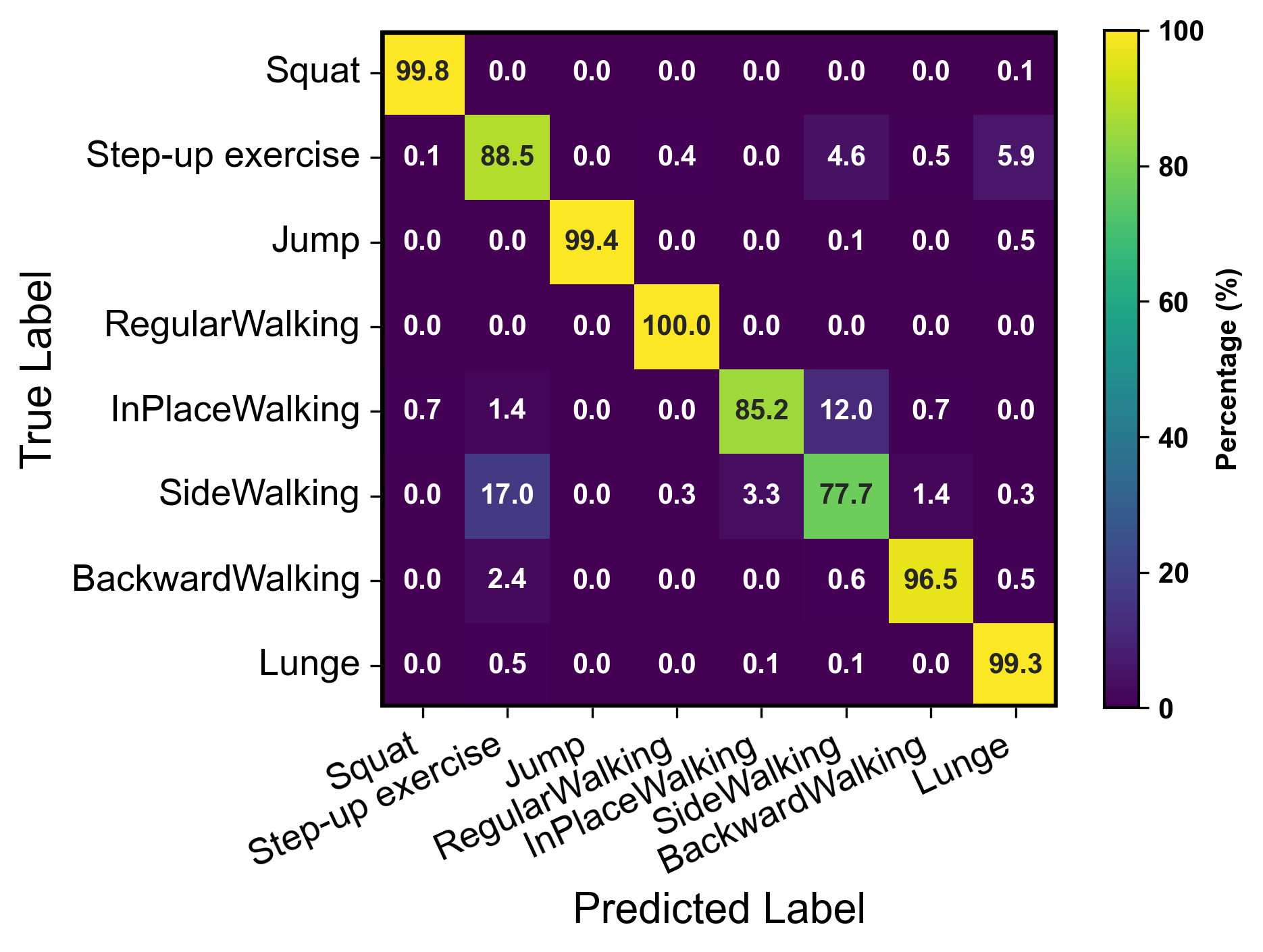}}
    
    \caption{\textbf{Confusion matrix for action classification.} While SCOTTI achieves high accuracy overall, misclassifications are observed between actions with similar tactile signal patterns. In particular, step-up exercises are sometimes misclassified as SideWalking or Lunge, and conversely, SideWalking samples are often confused with step-up exercises.}
    \label{fig:action_confusion_matrix}
\end{figure}

The results comparing our model with baselines are presented in Table~\ref{table:main}. Overall, our SCOTTI model achieves superior performance, successfully generalize to unseen subjects. For pose estimation, SCOTTI achieves the lowest MPJPE of 96.63 mm. In action classification, SCOTTI achieves an accuracy approximately 5.5 percentage points higher than STAT. For progress prediction, SCOTTI achieves an MSE of 0.0223, which is significantly lower than the random prediction benchmark of 0.1448. SCOTTI also highly outperforms random prediction in APP, with a value of 0.8952. The most notable aspect is that, while existing baselines are specialized for individual tasks, our SCOTTI model achieves these results more efficiently with a single unified model, and further introduces progress prediction as a novel task not addressed in prior work. 

We also compared our model with multi-task learning versions of the baselines, as shown in Table~\ref{table:multi}. Even when compared to multi-task versions of the baseline models, SCOTTI demonstrates superior performance. Comparing the single-task baselines in Table~\ref{table:main} with the multi-task baselines in Table~\ref{table:multi} reveals negligible performance improvements for the existing baselines. For instance, the MPJPE of the Voxel-CNN model only slightly improves from 104.91 mm in the single-task version to 103.55 mm in the multi-task version. Similarly, other models show either marginal improvements or slight performance degradation: Sep-CNN (100.57 mm $\rightarrow$ 101.05 mm), Res-CNN (98.51 mm $\rightarrow$ 99.00 mm), and GCN-Transformer (117.81 mm $\rightarrow$ 124.27 mm). For action classification, similar trends are observed. STAT improves slightly from 84.56\% $\rightarrow$ 86.36\%. 
These results supports that SCOTTI can effectively leverage the mutual benefits of the three tasks---pose estimation, action classification, and progress prediction.

\begin{table}[t]
    \caption{\textbf{Comparison of SCOTTI with single-task versions.}}
    \label{table:ablation}
    \centering
    \resizebox{\columnwidth}{!}{
    \begin{tabular}{l cc c cc}
        \toprule
        \multirow{2}{*}{\textbf{Model}} 
          & \begin{tabular}[c]{@{}c@{}} \textbf{Pose} \\ \textbf{Estimation} \end{tabular}
          & \begin{tabular}[c]{@{}c@{}} \textbf{Action} \\ \textbf{Classification} \end{tabular}
          & \multicolumn{2}{c}{\begin{tabular}[c]{@{}c@{}} \textbf{Progress} \\ \textbf{Prediction} \end{tabular}} \\
        \cmidrule(r){2-3} \cmidrule(l){4-4} \cmidrule(l){5-6}
         & \textbf{MPJPE (mm)} $\downarrow$
         & \textbf{Accuracy (\%)} $\uparrow$ 
         & \textbf{APP} $\uparrow$ & \textbf{MSE} $\downarrow$ \\
        \midrule
        Pose Only & 101.61 & - & - & -  \\
        Action Only & - & 87.39 & -  & - \\
        Progress Only & - & - & 0.8866 & 0.0264 \\
        \midrule
        \textbf{All Tasks} & \textbf{96.63} & \textbf{90.06} & \textbf{0.8952} & \textbf{0.0223} \\
        \bottomrule
    \end{tabular}
    }
\end{table}

\begin{figure*}[t]
    \centering
    \begin{subfigure}[t]{0.17\textwidth} 
        \centering
        \includegraphics[width=\linewidth]{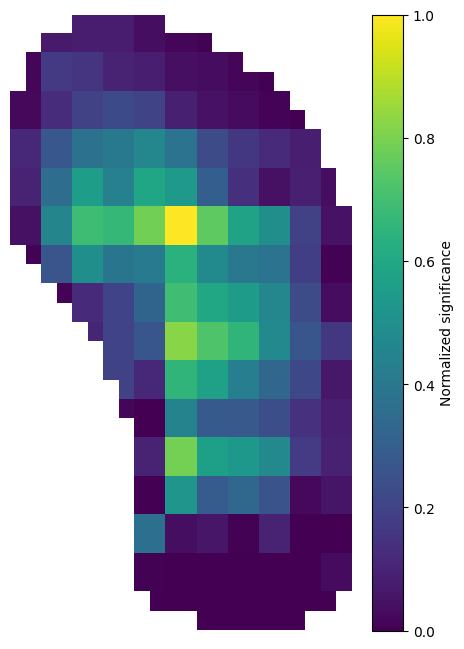}
        \caption{Importance of foot regions.}
    \end{subfigure}
    \begin{subfigure}[t]{0.66\textwidth} 
        \centering
        \includegraphics[width=\linewidth]{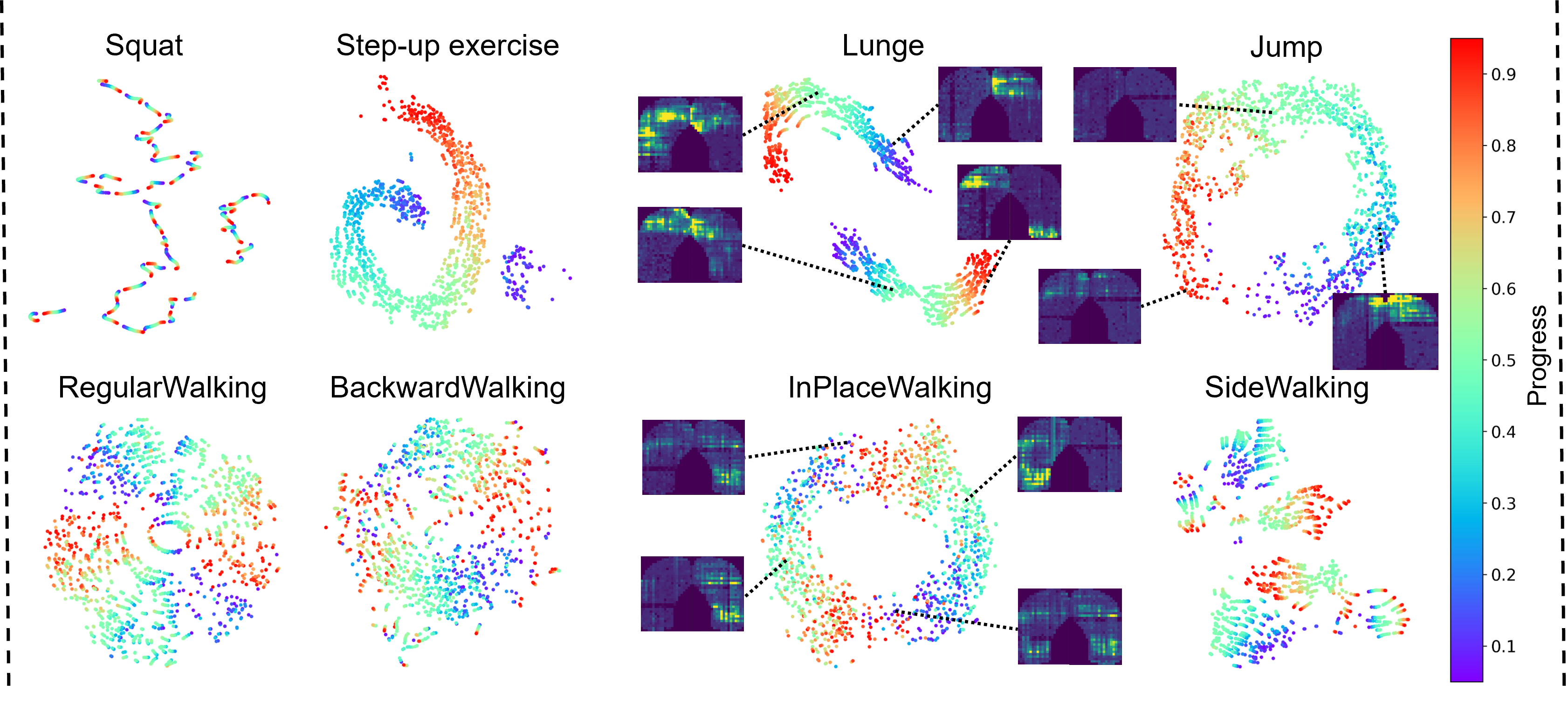}
        \caption{t-SNE by progress values.}
    \end{subfigure}
    \begin{subfigure}[t]{0.14\textwidth} 
        \centering
        \includegraphics[width=\linewidth]{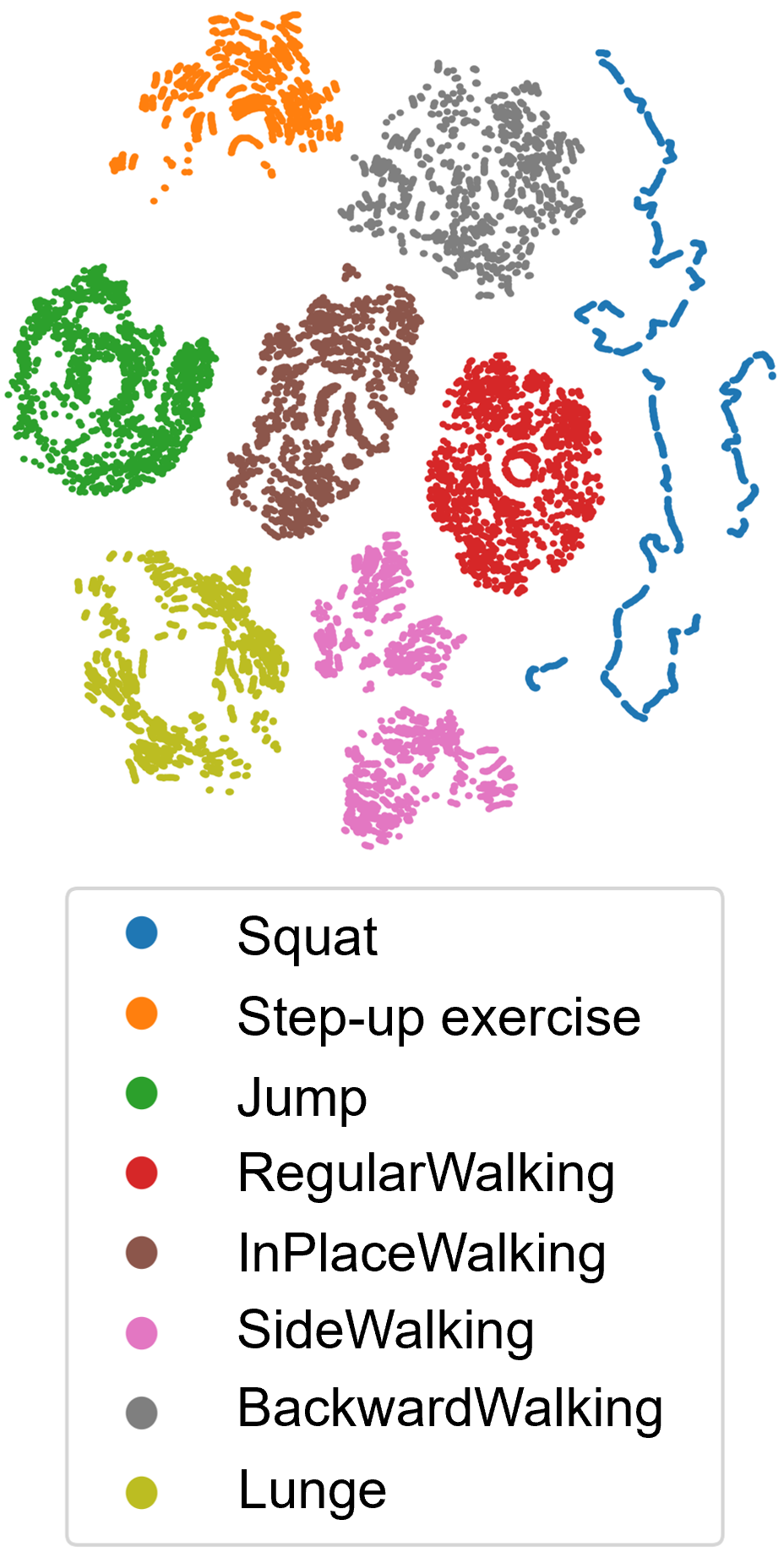}
        \caption{t-SNE by action labels.}
    \end{subfigure}
    \caption{\textbf{Analysis of SCOTTI.} \textbf{(a)} Importance of different foot regions for SCOTTI across tasks, highlighting the central foot regions' significance. \textbf{(b)} t-SNE visualization of shared features based on progress values, showing structured patterns for different actions. \textbf{(c)} t-SNE visualization of shared features based on action labels, demonstrating well-clustered features for individual actions.}
    \label{fig:sensing_and_t_sne}
\end{figure*}

We visualized the representative trials of pose estimation and progress prediction in Fig.~\ref{fig:qualitative}. The figure on the left illustrates the results for lunges, while the figure on the right presents the results for squats. These visualizations demonstrate that our model can accurately perform both pose estimation and progress prediction.

We also report the progress-margin curve, shown in Fig.~\ref{fig:progress_curve}, which shows that our model outperforms baselines on all margin values. 

The confusion matrix for action classification is visualized in Fig.~\ref{fig:action_confusion_matrix}. The results show that the model generally performs accurate classification; however, it occasionally fails to distinguish between actions with similar tactile signals. Notably, the misclassified samples often involve actions such as step-up exercises, side walking, and backward walking, which share similar tactile signal patterns.

\subsection{Ablation Study on Shared Representation}

For comparison, we constructed single-task variants of SCOTTI by retaining only the task-specific MLP head corresponding to each task, while keeping the encoder unchanged. We then demonstrate the benefits of multi-task learning by comparing the multi-task and single-task versions. The results are shown in Table~\ref{table:ablation}. In most metrics, multi-task model exhibits significant performance improvements. For instance, the MPJPE is reduced by 4.98 mm (101.61 $\rightarrow$ 96.63). The accuracy of action classification increases by 2.67\%p (87.39 $\rightarrow$ 90.06). For progress prediction, both APP and MSE show improvements, with lower error values in the multi-task model.

\subsection{Ablation Study on Sensing Region}

We visualized the importance of each foot region across all three tasks in Fig.~\ref{fig:sensing_and_t_sne}(a) to better understand human motion. The importance was calculated by masking each foot region of the data and observing the resulting increase in the total loss of the trained model. The results show that the central regions of the foot, where contact is more consistent, have higher importance values. This is because the edges of the foot are less likely to make contact and provide meaningful information. Interestingly, the central part of the foot is significantly more important than the heels. These findings enhance our understanding of human movement.

\subsection{Visualization of Shared Feature Space}

We show that our model has learned meaningful shared features from tactile signals by visualizing the shared features using t-SNE (t-distributed Stochastic Neighbor Embedding) \cite{van2008visualizing}. t-SNE is a nonlinear dimensionality reduction technique that projects high-dimensional data into a low-dimensional space while preserving local structure for visualization. The results are shown in the Fig.~\ref{fig:sensing_and_t_sne}(b) and (c). 

The middle figure visualizes the features based on progress values. Interestingly, walking types and lunges, which involve alternating foot movements, form ring-like structures. Along these rings, the progress increases from 0 to 1, with the two segments of the ring corresponding to each foot—one segment represents the left foot, and the other represents the right foot. For example, the left segment of in-place walking represents the scenario where the left foot is lifted while the right foot is in contact with the ground. Conversely, the right segment represents the scenario where the left foot is on the ground and the right foot is lifted. This can be observed through the tactile images shown in the figure. However, the features of jumping form a single segment, as both feet move simultaneously during a jump. For step-up exercises, the movement involves stepping onto the platform and stepping down, resulting in a continuous line with distinct start and end points. The right figure visualizes the features based on actions. The results show that features from the same actions cluster well. These findings indicate that the model has learned representations containing useful information of tactile signals, which can potentially be utilized for new tasks.
\section{Conclusion}

This paper presents SCOTTI, a Shared COnvolutional Transformer for Tactile Inference, designed to predict 3D human poses, action classes, and action progress simultaneously. By leveraging a shared representation, SCOTTI demonstrates that multi-task learning can outperform individual task-specific approaches, achieving superior results in all evaluated metrics and successfully generalizing on unseen subjects. Our contributions include the introduction of a unified model, the novel application of tactile signals for action progress prediction, and the creation of a comprehensive dataset with over 200,000 tactile and visual frames across diverse activities.

Experimental results highlight SCOTTI's capability to handle the complexity of multi-task learning efficiently. Additionally, the dataset and insights into the importance of specific foot regions and shared feature spaces underline the broader applicability of tactile-based research. These findings can aid leveraging tactile sensing in healthcare, sports analysis, robotics, and beyond.








\bibliographystyle{IEEEtran}
\bibliography{IEEEabrv, reference}

\end{document}